\documentclass{article} % For LaTeX2e
\usepackage{nips13submit_e,times}
\usepackage{hyperref}
\usepackage{url}
\usepackage[numbers]{natbib}
\usepackage{bm}
\usepackage{graphicx}
\usepackage[utf8]{inputenc}

\newcommand{\xx}{{\bf x}}
\newcommand{\zz}{{\bf z}}
\newcommand{\yy}{{\bf y}}
\newcommand{\vv}{{\bf v}}
\newcommand{\hh}{{\bf h}}

\newcommand{\bfl}{{\bf l}}
\newcommand{\WW}{{\bf W}}
\newcommand{\VV}{{\bf V}}

\newcommand{\bb}{{\bf b}}
\newcommand{\cc}{{\bf c}}
\newcommand{\bphi}{{\bm \phi}}
\newcommand{\ppi}{{\bm \pi}}

\newcommand{\sigm}{\mathrm{sigm}}

\title{Learning Multilingual Word Representations using a Bag-of-Words Autoencoder}

\author{{\bf Stanislas Lauly}\\
D\'epartement d'informatique\\
Universit\'e de Sherbrooke\\
{\it stanislas.lauly@usherbrooke.ca}\\
\And
 {\bf Alex Boulanger}\\
D\'epartement d'informatique\\
Universit\'e de Sherbrooke\\
{\it alex.boulanger@usherbrooke.ca}\\
\And
 {\bf Hugo Larochelle}\\
D\'epartement d'informatique\\
Universit\'e de Sherbrooke\\
{\it hugo.larochelle@usherbrooke.ca}}

\nipsfinalcopy 

\begin{document}

\maketitle

\begin{abstract}
Recent work on learning multilingual word representations usually
relies on the use of word-level alignements (e.g.\ infered with the
help of GIZA++) between translated sentences, in order to align the
word embeddings in different languages.  In this workshop paper, we
investigate an autoencoder model for learning multilingual word
representations that does without such word-level alignements. The
autoencoder is trained to reconstruct the bag-of-word representation
of given sentence from an encoded representation extracted from its
translation. We evaluate our approach on a multilingual document
classification task, where labeled data is available only for one
language (e.g.\ English) while classification must be performed in a
different language (e.g.\ French). In our experiments, we observe that
our method compares favorably with a previously proposed method that
exploits word-level alignments to learn word representations.
\end{abstract}

%%% TODO
% - j'aime pas p(v_i | \bphi)
% - dire à quelque part qu'on sait que normalement, h(phi(v)) 
%   est la représentation, mais qu'on décrit ça comme ça parce que phi(v) marche mieux en pratique

\section{Introduction}

Vectorial word representations have proven useful for multiple NLP
tasks~\cite{Turian+Ratinov+Bengio-2010,CollobertR2011}. It's been shown that meaningful representations, capturing
syntactic and semantic similarity, can be learned from unlabled
data. Along with a (usually smaller) set of labeled data, these
representations allows to exploit unlabeled data and improve
the generalization performance on some given task, even allowing to
generalize out of the vocabulary observed in the labeled data only.

While the majority of previous work has concentrated on the
monolingual case, recent work has started looking at learning word
representations that are aligned across languages~\cite{KlementievA2012,ZhouW2013,MikolovT2013}. These
representations have been applied to a variety of problems, including
cross-lingual document classification~\cite{KlementievA2012} and phrase-based
machine translation~\cite{ZhouW2013}. A common property of these approaches
is that a word-level alignment of translated sentences is leveraged,
either to derive a regularization term relating word embeddings
across languages~\cite{KlementievA2012,ZhouW2013}.

In this workshop paper, we experiment with a method to learn
multilingual word representations that does without word-to-word
alignment of bilingual corpora during training. We only require
aligned sentences and do not exploit word-level alignments
(e.g.\ extracted using GIZA++, as is usual). To do so, we propose a
multilingual autoencoder model, that learns to relate the hidden
representation of paired bag-of-words sentences.

We use these representations in the context of cross-lingual document
classification where labeled dataset can be available in one language,
but not in another one. With the multilingual word representations, we
want to learn a classifier with documents in one language and then use
it on documents in another language. Our preliminary experiments
suggest that our method is competitive with the representations
learned by \cite{KlementievA2012}, which rely on word-level alignments.

In Section~\ref{sec:autoencoder}, we describe the initial autoencoder
model that can learn a representation from which an input bag-of-words
can be reconstructed. Then, in Section~\ref{sec:multilingual_autoencoder},
we extend this autoencoder for the multilingual setting.
Related work is discussed in Section~\ref{sec:related_work} and
experiments are presented in Section~\ref{sec:experiments}.

\section{Autoencoder for Bags-of-Words}
\label{sec:autoencoder}

Let $\xx$ be the bag-of-words representation of a
sentence. Specifically, each $x_i$ is a word index from a fixed
vocabulary of $V$ words. As this is a bag-of-words, the order of the
words within $\xx$ does not correspond to the word order in the
original sentence. We wish to learn a $D$-dimensional vectorial
representation of our words from a training set of sentence
bag-of-words $\{\xx^{(t)}\}_{t=1}^T$.

We propose to achieve this by using an autoencoder model that encodes
an input bag-of-words $\xx$ as the sum of its word representations
(embeddings) and, using a non-linear decoder, is trained to reproduce
the original bag-of-words.

Specifically, let matrix $\WW$ be the $D\times V$ matrix whose columns are the
vector representations for each word. The aggregated representation
for a given bag-of-words will then be:
\begin{equation}
  \bphi(\xx) = \sum_{i=1}^{|\xx|} \WW_{\cdot,x_i}~~.
\end{equation}
To learn meaningful word representations, we wish to encourage
$\bphi(\xx)$ to contain information that allows for the reconstruction
of the original bag-of-words $\xx$. This is done by choosing a
reconstruction loss and by designing a parametrized decoder which will
be trained jointly with the word representations $\WW$ so as to
minimize this loss. 

Because words are implicitly high-dimensional objects, care must be
taken in the choice of reconstruction loss and decoder for stochastic
gradient descent to be efficient. For instance, \citet{DauphinY2011} recently
designed an efficient algorithm for reconstructing binary bag-of-words
representations of documents, in which the input is a fixed size
vector where each element is associated with a word and is set to 1
only if the word appears at least once in the document. They use
importance sampling to avoid reconstructing the whole $V$-dimensional
input vector, which would be expensive.

In this work, we propose a different approach. We assume that, from
the decoder, we can obtain a probability distribution over any word
$\widehat{x}$ observed at the reconstruction output layer
$p(\widehat{x}|\bphi(\xx))$. Then, we treat the input bag-of-words as
a $|\xx|$-trials multinomial sample from that distribution and use as
the reconstruction loss its negative log-likelihood:
\begin{equation}
  \ell(\xx) = \sum_{i=1}^{|\vv|} -\log p(\widehat{x}=x_i|\bphi(\xx))~.
\end{equation}
We now must ensure that the decoder can compute $p(\widehat{x}=x_i|\bphi(\xx))$
efficiently from $\bphi(\xx)$. Specifically, we'd like to avoid a
procedure scaling linear with the vocabulary size $V$, since $V$ will
be very large in practice. This precludes any procedure that would
compute the numerator of $p(\widehat{x}=w|\bphi(\xx))$ for each possible word $w$
separetly and normalize so it sums to one.

We instead opt for an approach borrowed from the work on neural
network language models~\cite{Morin+al-2005,Mnih+Hinton-2009}. Specifically, we use a probabilistic
tree decomposition of $p(\widehat{x}=x_i|\bphi(\xx))$.  Let's assume each
word has been placed at the leaf of a binary tree.  We can then treat
the sampling of a word as a stochastic path from the root of the tree
to one of the leaf. 

We note as $\bfl(x)$ as the sequence of
internal nodes in the path from the root to a given word $x$, with
$l(x)_1$ always corresponding to the root. We will not as
$\ppi(x)$ the vector of associated left/right branching choices on
that path, where $\pi(x)_k=0$ means the path branches left at internal
node $l(x)_k$ and branches right if $\pi(x)_k=1$ otherwise.  Then, the
probability $p(\widehat{x}|\bphi(\xx))$ of a certain word $x$ observed in the
bag-of-words is computed as
\begin{equation}
  p(\widehat{x}|\bphi(\xx)) = \prod_{k=1}^{|\ppi(\hat{x})|} p(\pi(\widehat{x})_k|\bphi(\xx))
\end{equation}
where $p(\pi(\widehat{x})_k|\bphi(\xx))$ is output by the decoder. By using a full
binary tree of words, the number of different decoder outputs required
to compute $p(\widehat{x}|\bphi(\xx))$ will be logarithmic in the vocabular size
$V$. Since there are $|\xx|$ words in the bag-of-words, at most
$O(|\xx| \log V)$ outputs are thus required from the decoder. This is
of course a worse case scenario, since words will share internal nodes
between their paths, for which the decoder output can be computed just
once. As for organizing words into a tree, as in \citet{LarochelleH2012} we used a
random assignment of words to the leaves of the full binary tree,
which we have found to work well in practice.

Finally, we need to choose of parametrized form for the decoder.
We choose the following non-linear form:
\begin{equation}
  p(\pi(\widehat{x})_k=1|\bphi(\xx)) = \sigm(b_{l(\hat{x}_i)_k} + \VV_{l(\hat{x}_i)_k,\cdot} \hh(\cc + \bphi(\xx)))
\end{equation}
where $\hh(\cdot)$ is an element wise non-linearity, $\cc$ is a
$D$-dimensional bias vector, $\bb$ is a ($V$-1)-dimensional bias
vector, $\VV$ is a $(V-1)\times D$ matrix and $\sigm(a) = 1 /
(1+\exp(-a))$ is the Sigmoid non-linearity. Each left/right
branching probability is thus modeled with a logistic regression model
applied on the non-linearly transformed representation of the input
bag-of-words $\bphi(\xx)$\footnote{While the literature
  on autoencoders usually refers to the post-nonlinearity activation
  vector as the hidden layer, we use a different description here simply to
  be consistent with the representation we will use for documents in our experiments,
  where the non-linearity will not be used}.

%(Say why we favor this reconstruction loss and decoder. 1. we can
%compute the loss and keep track of learning, and 2. we don't only want
%to model the presence of a word, but also its frequency count)

\section{Multilingual Bag-of-words}
\label{sec:multilingual_autoencoder}

\begin{figure}
\begin{center}
\includegraphics[width=0.5\textwidth]{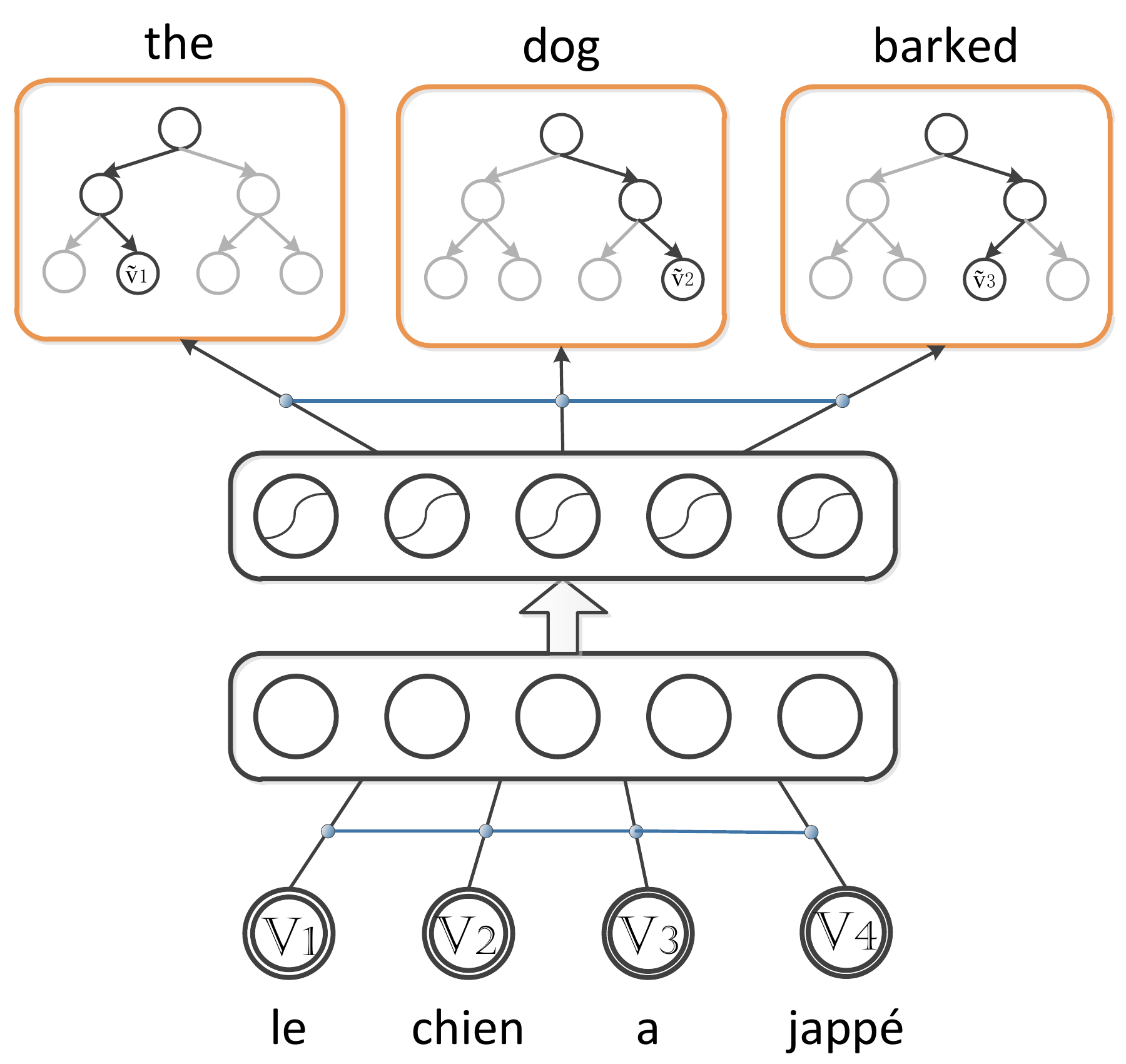}
\end{center}
\caption{Illustration of a bilingual autoencoder that learns to construct
the bag-of-word of the English sentence ``{\it the dog barked}''
from its French translation ``{\it le chien a jappé}''. The
horizontal blue line across the input-to-hidden connections
highlights the fact that these connections share the same parameters
(similarly for the hidden-to-output connections).}
\label{fig:bilingual_autoencoder}
\end{figure}

Let's now assume that for each sentence bag-of-words $\xx$ in some
source language ${\cal X}$, we have an associated bag-of-words $\yy$ for the same
sentence translated in some target language ${\cal Y}$ by a human expert. 
Assuming we have a training set of such $(\xx,\yy)$ pairs, we'd like
to use it to learn representations in both languages that are aligned,
such that pairs of translated words have similar representations.

To achieve this, we propose to augment the regular autoencoder
proposed in Section~\ref{sec:autoencoder} so that, from the sentence
representation in a given language, a reconstruction can be attempted
of the original sentence in the other language. 

Specifically, we now define language specific word representation
matrices $\WW^{x}$ and $\WW^{y}$, corresponding to the languages
of the words in $\xx$ and $\yy$ respectively. Let $V^x$ and 
$V^y$ also be the number of words in the vocabulary of both languages,
which can be different. The word representations however are of
the same size $D$ in both languages. The sentence-level representation
extracted by the encoder becomes
\begin{eqnarray}
  \bphi(\xx) = \sum_{i=1}^{|\xx|} \WW^{x}_{\cdot,x_i}~, & &   \bphi(\yy) = \sum_{i=1}^{|\yy|} \WW^{y}_{\cdot,y_i}~~.
\end{eqnarray}
From the sentence in either languages, we want to be able to perform a
reconstruction of the original sentence in any of the languages. In
particular, given a representation in any language, we'd like a
decoder that can perfrom a reconstruction in language ${\cal X}$ and
another decoder that can reconstruct in language ${\cal Y}$.  Again,
we use decoders of the form proposed in Section~\ref{sec:autoencoder},
but let the decoders of each language have their own parameters
$(\bb^x,\VV^x)$ and $(\bb^y,\VV^y)$:

\begin{eqnarray*}
  p(\widehat{x}|\bphi(\zz)) = \prod_{k=1}^{|\ppi(\hat{x})|} p(\pi(\widehat{x})_k|\bphi(\zz)),~&  p(\pi(\widehat{x})_k=1|\bphi(\zz)) = \sigm(b^x_{l(\hat{x}_i)_k} + \VV^x_{l(\hat{x}_i)_k,\cdot} \hh(\cc + \bphi(\zz)))\\
  p(\widehat{y}|\bphi(\zz)) = \prod_{k=1}^{|\ppi(\hat{y})|} p(\pi(\widehat{y})_k|\bphi(\zz)),~&  p(\pi(\widehat{y})_k=1|\bphi(\zz)) = \sigm(b^y_{l(\hat{x}_i)_k} + \VV^y_{l(\hat{y}_i)_k,\cdot} \hh(\cc + \bphi(\zz)))
\end{eqnarray*}
where $\zz$ can be either $\xx$ or $\yy$.  Notice that we share the
bias $\cc$ in the nonlinearity $\hh(\cdot)$ across decoders.  

This encoder/decoder structure allows us to learn a mapping within
each language and across the languages. Specifically, for a given pair
$(\xx,\yy)$, we can train the model to (1) construct $\yy$ from $\xx$,
(2) construct $\xx$ from $\yy$, (3) reconstruct $\xx$ from itself and
(4) reconstruct $\yy$ from itself.  In our experiments, performed each
of these 4 tasks simultaneously, combining the equally weighting the
learning gradient from each.  Experiments on various weighting schemes
should be investigated and are left for future work. Another
promising direction of investigation to the exploit the fact that
tasks (3) and (4) could be performed on extra monolingual corpora,
which is more plentiful.

\section{Related work}
\label{sec:related_work}

We mentioned that recent work has considered the problem of learning
multilingual representations of words and usually relies on word-level
alignments. \citet{KlementievA2012} propose to train simultaneously two neural
network languages models, along with a regularization term that
encourages pairs of frequently aligned words to have similar word
embeddings.  \citet{ZhouW2013} use a similar approach, with a different form
for the regularizor and neural network language models as
in~\cite{CollobertR2011}. In our work, we specifically investigate whether
a method that does not rely on word-level alignments can
learn comparably useful multilingual embeddings in the
context of document classification.

Looking more generally at neural networks that learn multilingual
representations of words or phrases, we mention the work of \citet{GaoJ2013}
which showed that a useful linear mapping between {\it separately
  training} monolingual skip-gram language models could be learned. 
They too however rely on the specification of pairs of words
in the two languages to align. \citet{MikolovT2013} also propose a method
for training a neural network to learn useful representations
of phrases (i.e.\ short segments of words), in the context of
a phrase-based translation model. In this case, phrase-level
alignments (usually extracted from word-level alignments)
are required.

\section{Experiments}\label{sec:experiments}

To evaluate the quality of the word embeddigns learned by our model,
we experiment with a task of cross-lingual document classication.  The
setup is as follows. A labeled data set of documents in some language
${\cal X}$ is available to train a classifier, however we are
interested in classifying documents in a different language ${\cal Y}$
at test time. To achieve this, we leverage some bilingual corpora,
which importantly is not labeled with any document-level categories.
This bilingual corpora is used instead to learn document representations
in both languages ${\cal X}$ and ${\cal Y}$ that are enroucaged
to be invariant to translations from one language to another.
The hope is thus that we can successfully apply the classifier trained on
document representations for language ${\cal X}$ directly
to the document representations for language ${\cal Y}$.

\subsection{Data}

We trained our multilingual autoencoder to learn bilingual word representation between
English and French and between English and German. To train the autoencoder,
we used the English/French and English/German section pairs of the Europarl-v7
dataset\footnote{\url{http://www.statmt.org/europarl/}}. 
This data is composed of about two million sentences,
where each sentence is translated in all the relevant languages. 
%The
%training, validation and testing sets were obtained using 70\%, 15\% and 15\%
%split of the Europarl-v7 dataset.

For our crosslingual classification problem, we used the English,
French and German sections of the Reuters RCV1/RCV2 corpus, as
provided by
\citet{AminiM2009}\footnote{\url{http://multilingreuters.iit.nrc.ca/ReutersMultiLingualMultiView.htm}}.
There are 18758, 26648 and 29953 documents (news stories) 
for English, French and German respectively.
Document categories are organized in a hierarchy in this dataset.  A
4-category classification problem was thus created by using the 4
top-level categories in the hierarchy (CCAT, ECAT, GCAT and MCAT).
The set of documents for each language is split into training,
validation and testing sets of size 70\%, 15\% and 15\%
respectively. The raw documents are represented in the form of a
bag-of-words using a TFIDF-based weighting scheme.  Generally, this
setup follows the one used by \citet{KlementievA2012}, but uses the preprocessing
pipeline of \citet{AminiM2009}.

\subsection{Crosslingual classification}

As described earlier, crosslingual document classification
is performed by training a document classifier on documents in
one language and applying that classifier on documents in
a different language at test time. Documents in a language are representated
as a linear combination of its word embeddings learned for that language.
Thus, classification performance relies heavily on the quality of 
multlingual word embeddigns between languages, and specifically
on whether similar words across languages have similar embeddings.

Overall, the experimental procedure is as follows. 
\begin{enumerate}
  \item Train bilingual word representations $\WW^{x}$ and $\WW^{y}$
    on sentence pairs extracted from Europarl-v7 for languages
    ${\cal X}$ and ${\cal Y}$ (we use a separate validation set
    to early-stop training).
  \item Train document classifier on the Reuters training
    set for language ${\cal X}$, where documents
    are represented using the word representations $\WW^{x}$ 
    (we use the validation
    set for the same language to perform model selection).
  \item Use the classifier trained in the previous step
    on the Reuters test set for language ${\cal Y}$, using
    the word representations $\WW^{y}$ to represent the
    documents.
\end{enumerate}
We used a linear SVM as our classifier. 

We compare our representations to those learned by
\citet{KlementievA2012}\footnote{The trained word embeddings were downloaded from
  \url{http://people.mmci.uni-saarland.de/~aklement/data/distrib/}}.
This is achieved by simply skipping the first step of training the
bilingual word representations and directly using those of \citet{KlementievA2012}
in step 2 and 3.  The provided word embeddings are of size 80 for the
English and French language pair, and of size 40 for the English and
German pair.  The vocabulary used by \citet{KlementievA2012} consisted in 43614
words in English, 35891 words in French and 50110 words in German. The
same vocabulary was used by our model, to represent the Reuters
documents.

In all cases, document representations were obtained by 
multiplying the word embeddings matrix with either
the TFIDF-based bag-of-words feature vector or its binary
version (the choice of which method to use is made based on the validation
set performance).  We normalized the TFIDF-based weights to sum to one. 

%We tried a binary
%bag-of-words because our model was originally trained with pair of
%sentence where each words where rarely repeat more than once. We found
%that better validation results where obtain for our model with binary
%bag-of-words. We tried the same thing for the Klementiev
%representation, but the best result for him was with the TFIDF
%bag-of-words.

Test set classification error results are reported in
Table~\ref{tab:results}.  We observe that
the word representations learned by our autoencoder are competitive
with those provided by~\citet{KlementievA2012}. One will notice that the results for
\citet{KlementievA2012} are worse than those reported in the original
reference. This difference might be due to the fact that our preprocessing
of the Reuters data, which comes from \citet{AminiM2009}, is different from the
one in \citet{KlementievA2012}. In particular, we note that \citet{KlementievA2012} ignored
documents that belonged to multiple categories, while \citet{AminiM2009}
included them by assigning them to the category with the least
training examples. 

\begin{figure}[h]
\begin{center}
\includegraphics[angle=90,width=0.9\textwidth]{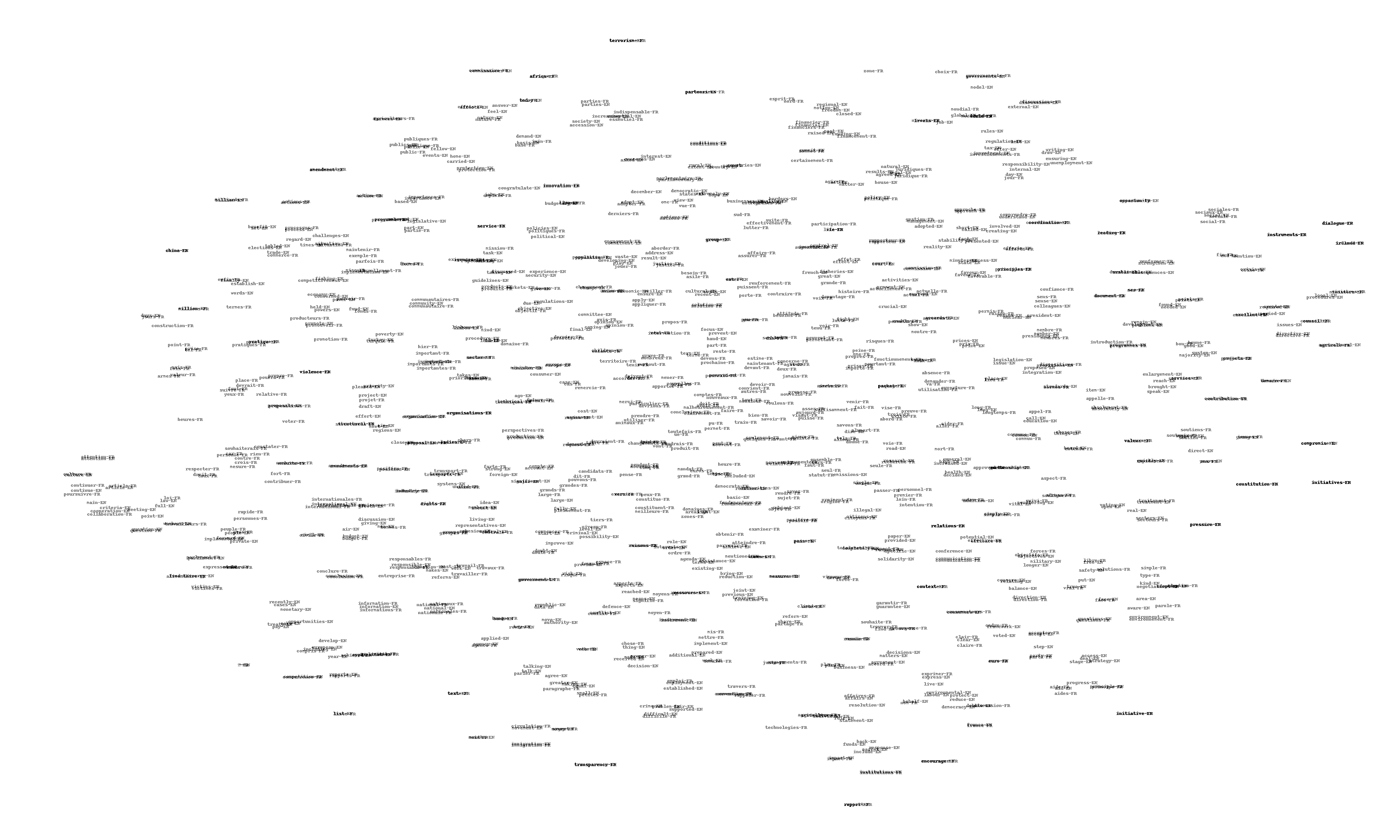}
\end{center}
\caption{A t-SNE 2D visualization of the learned English/French word representations
(better visualized on a computer). Words hyphenated with ``EN'' and ``FR'' are English and
French words respectively.}
\label{graphTSNE}
\end{figure}

Table~\ref{tab:results} also shows, for a few French words, what are
the nearest neighbor words in the English embedding space.  A more
complete picture is presented in the t-SNE visualization~\cite{VanDerMaaten08} of
Figure~\ref{graphTSNE}.  It shows a 2D visualization of the
French/English word embeddings, for the 600 most frequent words in
both languages.  Both illustrations confirm that the multilingual
autoencoder was able to learn similar embeddings for similar words
across the two languages.

\begin{table}[t]
\begin{center}
\begin{minipage}{.5\textwidth}
\begin{tabular}{|l|l|l|}
\hline
& Train FR /  &  Train EN /\\
 & Test EN & Test FR \\
\hline
 Klementiev  & 34.9\% & 49.2\% \\
\hline
Our embeddings  & {\bf 27.7\%} & {\bf 32.4\%} \\
\hline
\hline
{\bf } & Train GR / & Train EN / \\
 & Test EN  & Test GR \\
\hline
Klementiev et al.\  & 42.7\% & 59.5\% \\
\hline
Our embeddings  & {\bf 29.8\%} & {\bf 37.7\%} \\
\hline
\end{tabular}
\end{minipage}
~~~~~~~~\begin{minipage}{.29\textwidth}
\includegraphics[width=1\textwidth]{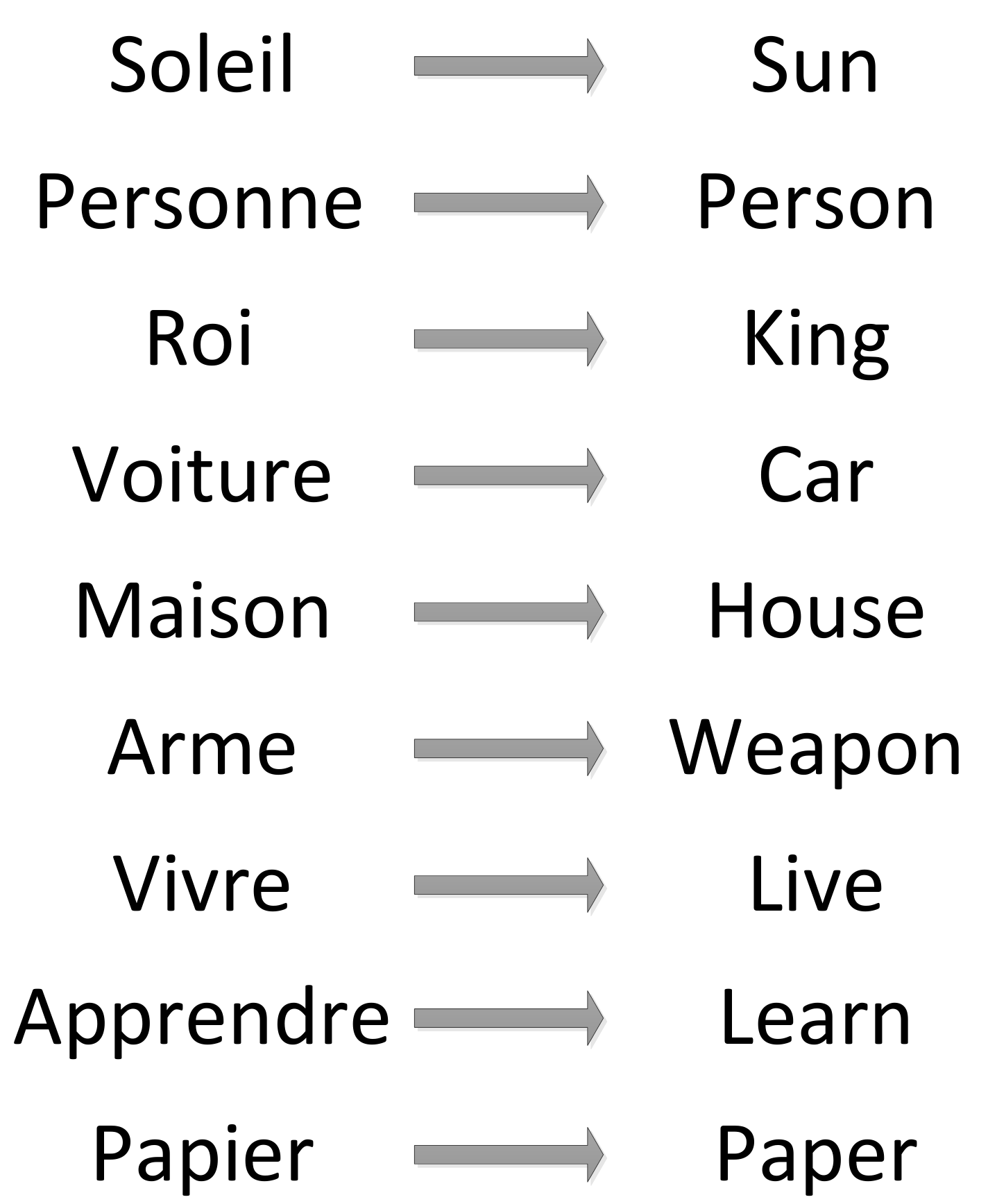}
\end{minipage}
\end{center}
\caption{{\bf Left:} Crosslingual classification error results, for  English/French
pair {\bf (Top)} and English/German pair {\bf (Bottom)}. {\bf Right:} For each French
word, its nearest neighbor in the English word embedding space.}
\label{tab:results}
\end{table}

%The figure \ref{graphClosestWords} show the nearest neighbor in
%English for a French word in the bilingual representation space
%learned by our model.

\section{Conclusion and Future Work}

We presented evidence that meaningful multilingual word
representations could be learned without relying on word-level
alignments. Our proposed multilingual autoencoder was able to perform
competitively on a crosslingual document classification task, compared
to a word representation learning method that exploits word-level
alignments. 

Encouraged by these preliminary results, our future work will
investigate extensions of our bag-of-words multilingual autoencoder to
bags-of-ngrams, where the model would also have to learn
representations for short phrases. Such a model should be particularly
useful in the context of a machine translation system. Thanks to the
use of a probabilistic tree in the output layer, our model could
efficiently assign scores to pairs of sentences. We thus think it could
act as a useful, complementary metric in a standard phrase-based translation system.

{\bf Acknowledgements}

Thanks to Alexandre Allauzen for his help on this subject. Thanks also to Cyril Goutte for providing the classification dataset. Finally, a big thanks to Alexandre Klementiev and Ivan Titov for their help.

\bibliographystyle{unsrtnat.bst}
\bibliography{multilingual_autoencoder}

\end{document}